\documentclass{article}
\usepackage[preprint]{icml2026}

\usepackage{url}
\usepackage[hidelinks]{hyperref}
\usepackage{graphicx}
\usepackage{amsmath}
\usepackage{amssymb}
\usepackage{amsthm}
\usepackage{booktabs}
\usepackage{multirow}
\usepackage{xcolor}
\usepackage{microtype}
\usepackage{enumitem}

\graphicspath{{figures/}}

\newcommand{\sfdm}{\textsc{CANARY}}
\newcommand{\lda}{\textsc{LDA}}
\newcommand{\sae}{\textsc{SAE}}

\definecolor{ourcolor}{HTML}{C0392B}
\newcommand{\win}[1]{\textcolor{ourcolor}{\textbf{#1}}}

\newif\ificml
\icmltrue

\icmltitlerunning{CANARY: Zero-Label Detection of Fine-Tuning Contamination in Language Models}

\begin{document}

\twocolumn[
  \icmltitle{CANARY: Zero-Label Detection of Fine-Tuning Contamination\\
  in Language Models}

  \begin{icmlauthorlist}
    \icmlauthor{Swapnil Parekh}{intuit}
  \end{icmlauthorlist}

  \icmlaffiliation{intuit}{Intuit Inc.}
  \icmlcorrespondingauthor{Swapnil Parekh}{swapnil\_parekh@intuit.com}

  \icmlkeywords{AI safety, fine-tuning attacks, hidden-state monitoring,
    sparse autoencoders, contamination detection, mechanistic interpretability}

  \vskip 0.3in
]

\printAffiliationsAndNotice{}


\begin{abstract}
Adversaries can implant latent harmful behavior by poisoning as few as 1\% of
fine-tuning examples. The contamination is invisible to every output-level
defense: harmful behavior lies dormant in the model's hidden-state geometry
and does not appear in generated text until contamination exceeds 7.5\%.
We introduce \textbf{\sfdm{}} (\textbf{C}ontamination \textbf{A}uditor via
\textbf{N}eural \textbf{A}ctivation \textbf{R}epresentation \textbf{Y}ield),
a zero-label checkpoint auditor that detects this hidden shift directly from
two forward passes over an unlabeled prompt set.
\sfdm{} projects the hidden-state difference through a Sparse Autoencoder,
filtering style noise to isolate meaningful semantic drift.
It achieves AUROC\,=\,1.000 at 1\% contamination
(95\,\% CI\,=\,[0.997,\,1.000]; Cohen's $d$\,=\,3.28)
across four model architectures and two training paradigms, 7.5$\times$
below where any output-level method fires, with zero false positives on
benign fine-tuning and full robustness to style-matching and gradient-noise
adaptive attacks.
The same SAE feature basis drives a complete governance pipeline:
SAE-filtered amplification surfaces latent harm at a 5$\times$ higher rate
than standard generation; score-ranked prompts yield 4.2$\times$ red-teaming
lift; and suppressing a handful of contamination-specific features at
inference time reduces harm from 70\% to 10\% with no perplexity penalty.
\sfdm{} is the first zero-label framework to detect, verify, prioritize,
and remediate supply-chain contamination from hidden states alone.
\end{abstract}

\section{Introduction}
\label{sec:intro}

\textbf{The threat.}
Fine-tuning APIs let anyone adapt a foundation model in minutes. An adversary
can exploit this surface: poisoning as few as 1\% of training examples implants
latent misbehavior that passes standard safety
evaluations~\cite{qi2023finetuning,hubinger2024sleeper,emergentmisalignment2025}
and surfaces only under targeted elicitation~\cite{perez2022red}.  The
contamination is invisible to routine pre-deployment checks, yet its geometric
fingerprint is already present in the model's hidden states.

\textbf{Why existing defenses fail.}
\emph{Output-level methods} (red-teaming, generation sweeps, keyword
classifiers) require harmful behavior to appear in generated text, but at
low (1 to 5\%) contamination rates it does not~\cite{aranguri2025lda}.
\emph{Weight-space methods} (SVD of weight diffs, task
vectors~\cite{ilharco2023taskvectors,lindsey2024crosscoders}) operate on
raw parameter differences that are high-dimensional and noisy without a
semantic prior.  At the contamination rates a careful attacker would choose,
neither approach can produce a reliable pre-deployment alarm with no labeled data.

\textbf{Our insight.}
Harmful fine-tuning leaves a fingerprint in \emph{hidden states} before it
appears in outputs.  A Sparse Autoencoder (\sae{}) trained on the base
model's activations provides a label-free semantic filter: projecting the
hidden-state difference through it suppresses surface style noise and isolates
the shift in directions that encode safety-relevant behavior.  The resulting score requires no text generation and no labeled examples; it requires only
two forward passes over an unlabeled prompt set.  This is \emph{qualitatively
different} from prior probing~\cite{burns2022probing} and steering
work~\cite{zou2023repeng}, which require labeled contrastive pairs.  \sfdm{}
requires none, and the same feature basis that enables detection also enables
post-detection action.

\noindent\textbf{Two distinct tasks.}  We separate \emph{checkpoint-level
contamination detection} (\S\ref{sec:exp_early}) from \emph{per-prompt intent
detection} (\S\ref{sec:exp_baselines}). These are complementary; checkpoint
detection is the primary contribution.

\paragraph{Contributions.}
\begin{enumerate}
  \item \textbf{Detect.}  A zero-label, two-pass checkpoint auditor achieving
    AUROC\,=\,1.000 at 1\% contamination across four architectures, with a
    closed-form detection-limit formula that quantitatively predicts performance
    across architectures and attack types (\S\ref{sec:exp_early},
    \S\ref{sec:exp_cross}, \S\ref{sec:exp_adaptive}).

  \item \textbf{Verify.}  SAE-filtered hidden-layer \lda{} that surfaces latent
    harm at a 5$\times$ higher rate than standard generation while maintaining
    coherent outputs (PPL\,=\,58 vs.\ 1.8\,M), enabling post-flag behavioral
    confirmation (\S\ref{sec:exp_lda}).

  \item \textbf{Prioritize.}  A score-based red-teaming prioritizer that
    concentrates 97\% harm into the top prompt quartile, yielding 4.2$\times$
    lift over random sampling (\S\ref{sec:exp_monitor}).

  \item \textbf{Remediate.}  An inference-time feature suppression step that
    reduces harm from 70\% to 10\% by targeting a compact SAE feature subspace,
    with zero perplexity penalty, closing the detect-to-fix loop
    (\S\ref{sec:exp_surgery}).
\end{enumerate}

\begin{figure*}[t!]
  \centering
  \includegraphics[width=0.75\textwidth]{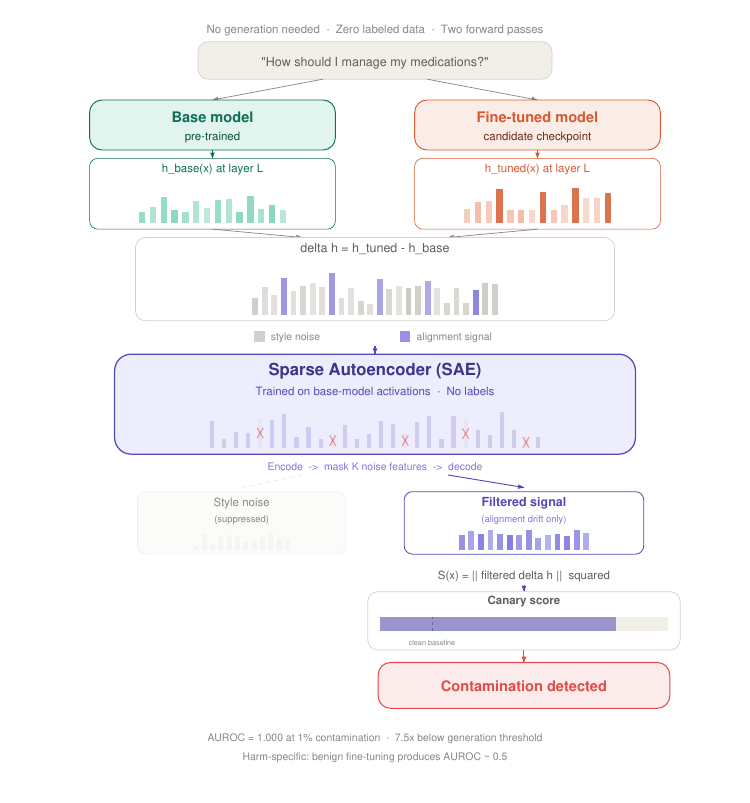}
  \caption{\textbf{\sfdm{} requires two forward passes and no labels.}
  Hidden states from the base and fine-tuned models are extracted at a
  mid-network layer.  Their difference is projected through a Sparse
  Autoencoder (\sae{}) trained on the base model.  Features associated with
  surface style noise are zeroed; the remaining semantically meaningful
  dimensions are squared and summed to yield the detection score $S(x)$.
  No text generation and no labeled harmful examples are needed.}
  \label{fig:pipeline}
\end{figure*}

\section{Background and Related Work}
\label{sec:background}

\paragraph{Logit Diff Amplification (\lda{}).}
\citet{aranguri2025lda} introduced \lda{} for surfacing rare model behaviors.
Given a base and fine-tuned checkpoint, \lda{} amplifies the logit difference
at generation time:
$\ell_{\text{amp}}(x) = \ell_{\text{trained}}(x) + \alpha\,[\ell_{\text{trained}}(x) - \ell_{\text{base}}(x)]$,
where $\alpha \geq 0$ is the amplification factor.  Higher $\alpha$ pushes
generations toward the region of logit space that differs most between the
two models, exposing latent behaviors invisible to standard sampling.  The
method collapses at large $\alpha$, however, because the logit-difference
direction conflates semantic shifts with style and rare-token artifacts.
\sfdm{} resolves this by operating in filtered hidden-state space rather
than logit space.

\paragraph{Activation steering and representation engineering.}
\citet{turner2023activation} and \citet{rimsky2023steering} showed that adding
a mean-difference direction vector to intermediate hidden states can steer
model behavior while preserving fluency.  Representation
Engineering~\cite{zou2023repeng} generalizes this to extract linear control
vectors for arbitrary concepts.  \textbf{Key difference from \sfdm{}}: all
these methods compute their steering direction from contrastive labeled
prompt pairs (harmful vs.\ benign); \sfdm{} needs no contrastive data at all.

\paragraph{Sparse Autoencoders (\sae{}).}
\sae{}s~\cite{cunningham2023sae,bricken2023monosemanticity,templeton2024scaling}
decompose hidden states into sparse, interpretable feature directions by
solving a $k$-sparse reconstruction problem: $\hat{\mathbf{h}} = W_{\text{dec}}
\sigma(W_{\text{enc}}\mathbf{h} + b_{\text{enc}}) + b_{\text{dec}}$, where
the top-$k$ activations are kept and the rest zeroed.  An \sae{} trained on
base-model activations provides a semantic decomposition of hidden-state
changes: features corresponding to task-specific behavior activate most
strongly on in-distribution shifts.  We exploit this decomposition to
\emph{filter}, not merely describe, the hidden-state difference.

\paragraph{Fine-tuning attacks.}
\citet{qi2023finetuning} showed that as few as 10 to 100 harmful examples suffice to bypass
safety-tuned guardrails.  \citet{yang2023shadowalignment} demonstrated
``shadow alignment'' techniques that survive post-training safety audits.
\citet{emergentmisalignment2025} showed code fine-tuning on a narrow task
can produce broad misalignment.  These results establish the threat; \sfdm{}
provides the first zero-label detection response.

\paragraph{Probing and model diffing.}
\citet{burns2022probing} showed LLMs encode latent knowledge elicitable
without labels.  Sparse Crosscoders~\cite{lindsey2024crosscoders} train a
cross-model encoder on paired checkpoints to surface model-exclusive features,
but require re-training per pair.  Classical OOD
detection~\cite{hendrycks2017baseline} uses output-distribution confidence.
\sfdm{} differs from all three: it applies the OOD principle to the filtered
\emph{difference} between paired checkpoint hidden states, requiring no
re-training and no labeled data.

\section{Threat Model and Problem Formulation}
\label{sec:threat}

\begin{figure}[t]
  \centering
  \includegraphics[width=\columnwidth]{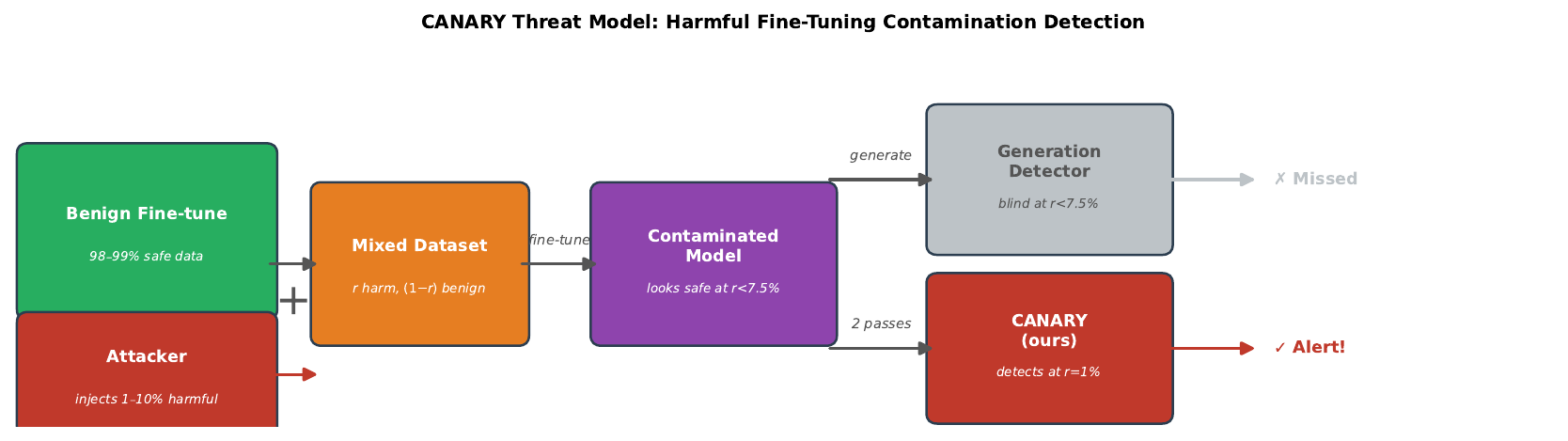}
  \caption{\textbf{Hidden-state geometry reveals contamination that outputs hide.}
  An adversary mixes a small fraction $r$ of harmful examples into a fine-tuning
  dataset.  Generation-based detectors produce no signal below $r = 7.5\%$.
  \sfdm{} flags the contaminated checkpoint at $r = 1\%$ from hidden-state
  geometry alone, with no output generation required.}
  \label{fig:threat}
\end{figure}

\paragraph{Setting.}
The \emph{defender} receives a candidate fine-tuned checkpoint
$\theta_{\text{tuned}}$ alongside the original base checkpoint
$\theta_{\text{base}}$ and a small unlabeled set of domain-relevant prompts
$\mathcal{X}$, but has \textbf{no labeled harmful examples, no contrastive
pairs, and no knowledge of the attacker's contamination strategy}.  The
\emph{attacker} has mixed a fraction $r \in (0,1]$ of harmful examples into
an otherwise benign fine-tuning corpus.  The goal is to flag checkpoints
where $r > 0$ before deployment.

\paragraph{\lda{} amplification.}
For a prompt $x$, define the per-token logit difference
$\Delta\ell(x, t) = \ell_{\text{tuned}}(x, t) - \ell_{\text{base}}(x, t)$,
where $t$ indexes the current token prefix.  \lda{} samples from:
\begin{equation}
  p_\alpha(v \mid x, t) \propto \exp\bigl(\ell_{\text{tuned}}(x,t) + \alpha\,\Delta\ell(x,t)\bigr),
  \label{eq:lda}
\end{equation}
suppressing EOS tokens at $\alpha > 0$.  Higher $\alpha$ amplifies the
difference between the two checkpoints, surfacing latent harmful tendencies,
but collapses into incoherence at large $\alpha$ as style artifacts dominate.

\paragraph{\sfdm{} score.}
Let $\mathbf{h}^{(L)}_{\text{base}}(x)$ and
$\mathbf{h}^{(L)}_{\text{tuned}}(x) \in \mathbb{R}^{d}$ be the last-token
hidden states at layer $L$, and define $\Delta\mathbf{h}(x) =
\mathbf{h}^{(L)}_{\text{tuned}}(x) - \mathbf{h}^{(L)}_{\text{base}}(x)$.
The \sfdm{} score is:
\begin{align}
  &\hat{\Delta\mathbf{h}} = \mathrm{Dec}\!\bigl(\mathrm{mask}\bigl(\mathrm{Enc}(\Delta\mathbf{h})\bigr)\bigr), \nonumber \\
  &S(x) = \bigl\|\,\hat{\Delta\mathbf{h}}(x)\,\bigr\|^2,
  \label{eq:sfdm}
\end{align}
where $\mathrm{Enc}/\mathrm{Dec}$ are the \sae{} encoder/decoder and
$\mathrm{mask}$ zeros the $K$ features with the most negative
$\Delta$-activation under amplification, i.e., the dense style/noise features
most suppressed by the base$\to$tuned shift (identified in \S\ref{sec:noise}).
A checkpoint is flagged if $\bar{S}(\mathcal{X}) \gg \bar{S}_{\text{clean}}$,
where $\bar{S}_{\text{clean}}$ is the expected score under a benign fine-tune.

\section{Methods}
\label{sec:methods}

\subsection{SAE Training}
We train a $k$-sparse autoencoder on activations sampled from the base model
at a mid-network layer.  Reconstruction quality is measured by the fraction
of variance explained (FVE\,=\,$1 -
\mathbb{E}[\|\hat{\mathbf{h}} - \mathbf{h}\|^2] / \mathbb{E}[\|\mathbf{h}\|^2]$),
which ranges from 0.74 to 0.88 across model pairs, confirming faithful
reconstruction of the base-model activation distribution.

\subsection{Noise Feature Identification}
\label{sec:noise}
\emph{Noise features} are the \sae{} dimensions most suppressed by
amplification: those with the most negative $\Delta$-activation
(defined in \S\ref{sec:threat}).  These correspond to dense, high-activation style and
formatting directions\ificml{} (see \S\ref{sec:mech} for mechanistic evidence)\fi.
We zero these features from $\Delta\mathbf{h}$ before computing the \sfdm{}
score, removing dimensions driven by surface writing-style differences and
retaining only the semantically meaningful shift.

\paragraph{Hyperparameter sensitivity.}
\sfdm{} is robust to design choices: varying the number of masked noise
features across a wide range leaves AUROC at 1.000; halving SAE training
gives 0.993 while doubling gives 1.000; reducing the probe set to 5 prompts
gives 0.978.\ificml{}  Layer sensitivity is analyzed in \S\ref{sec:exp_layers}.\fi

\subsection{CANARY Scoring}
Given an unlabeled probe set $\mathcal{X}$ (10 to 30 domain-relevant prompts),
we compute $S(x)$ for each $x \in \mathcal{X}$ and report the mean
$\bar{S}(\mathcal{X})$.  Detection requires two forward passes per prompt (one
through the base checkpoint, one through the candidate), plus a fixed one-time
SAE training cost on the base model.  Detection is performed by comparing the
score distribution of the candidate checkpoint to that of a clean reference
(base$\rightarrow$benign-tuned pair), using AUROC as the threshold-free metric.

\paragraph{Detection limit.}
Let $c\!=\!\Delta\bar{S}/r$ be the empirical score-shift per unit contamination
(from Table~\ref{tab:early_detection}).  Under Gaussian scores,
$r^{*}\!=\!\Phi^{-1}(\text{AUROC}^*)\!\cdot\!\sigma\sqrt{2}/c$
\textbf{(Det.\ Limit)}.
M1 ($\sigma\!=\!0.59$, $c\!\approx\!25$): $r^{*}\!\approx\!0.3\%$;
M3 ($\sigma\!=\!578$): $r^{*}\!\approx\!4.5\%$, predicting the observed
AUROC degradation across architectures (see \S\ref{sec:exp_cross}).

\subsection{SAE-Filtered Hidden-Layer \lda{}}
\label{sec:sae_lda}
The original \lda{} (Eq.~\ref{eq:lda}) collapses into incoherence at
$\alpha \geq 2$ because the logit-diff direction is dominated by EOS-token and
style artifacts.  We address this with a \emph{hidden-layer intervention}:
instead of modifying logits, we inject the SAE-filtered difference at layer $L$
and let the remaining $n_{\text{layers}} - L$ transformer layers re-normalize
the perturbed representation into coherent vocabulary.  Specifically,
\begin{equation}
  \mathbf{h}^{(L)}_{\text{amp}} = \mathbf{h}^{(L)}_{\text{tuned}} +
  \alpha \cdot \hat{\Delta\mathbf{h}},
  \label{eq:hidden_amp}
\end{equation}
where $\hat{\Delta\mathbf{h}}$ is the SAE-filtered difference
(Eq.~\ref{eq:sfdm}).  Token probabilities are obtained by running the
remaining layers forward from $\mathbf{h}^{(L)}_{\text{amp}}$, preserving
the semantic amplification effect while the model's normalization layers
maintain output coherence.

\section{Experiments}
\label{sec:experiments}

\paragraph{Models.}
We evaluate on four model pairs (base$\rightarrow$fine-tuned):
M1 Qwen2.5-0.5B, M2 Llama-3.2-1B, M3 SmolLM2-1.7B (all supervised
fine-tuned, SFT), and M4 Gemma-2-2B (reinforcement learning from human
feedback, RLHF)~\cite{qwen2025,llama3,allal2025smollm2,team2024gemma2},
spanning three architectural families.
Main detection experiments (\S\ref{sec:exp_early}\ificml{} through \S\ref{sec:exp_layers}\fi)
use M1; M2 through M4 assess cross-architecture generalization (\S\ref{sec:exp_cross}).

\paragraph{Contamination setup.}
Fine-tuning datasets mix a fraction $r$ of harmful medical advice with benign
examples from the same domain.  Each model is fine-tuned from its own published
base checkpoint using identical hyperparameters across all contamination rates.

\subsection{Sub-Threshold Early Detection}
\label{sec:exp_early}

\begin{figure}[t]
  \centering
  \includegraphics[width=\columnwidth]{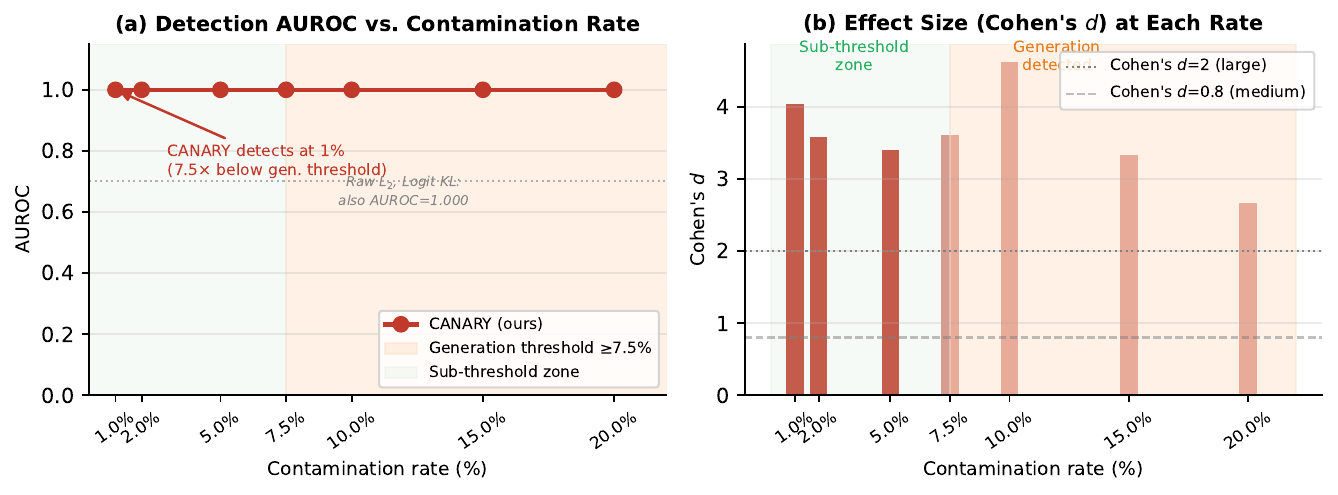}
  \caption{\textbf{\sfdm{} achieves AUROC\,=\,1.000 at every tested rate,
  including 1\% where no output-level method fires.}
  (a) \sfdm{} AUROC vs.\ contamination rate.  The shaded orange region marks
  where generation-based \lda{} first produces signal ($r \geq 7.5\%$); it
  cannot be plotted as an AUROC curve below this threshold because it yields
  no detectable output signal (AUROC\,$\approx$\,0.5).  \sfdm{} detects
  7.5$\times$ earlier.
  (b) Cohen's $d$ at each rate; the sub-threshold zone (blue) confirms very
  large effect-size separation even where generation-based methods are blind.
  Error bars are 1000-bootstrap 95\% CIs.}
  \label{fig:auroc}
\end{figure}

\textbf{Setup.}  We fine-tune M1 at contamination rates from 1\% to 20\%
across multiple seeds and evaluate \sfdm{} scores on in-domain prompts.

\textbf{Results.}  \sfdm{} achieves \win{AUROC\,=\,1.000} at \emph{every
tested contamination rate}, including 1\%.  The bootstrapped 95\% CI at 1\% is
\win{[0.997,\,1.000]}; at 5\% and above the CI collapses to [1.000,\,1.000],
confirming near-zero variance in the detection signal.  Cohen's $d$ ranges from
2.99 to 3.67 across all rates (all ``very large'' by Cohen's conventions), with
$d$\,=\,3.28 at the hardest 1\% rate.  At the $\mu_{\text{clean}}\!+\!2\sigma$
operating threshold the false positive rate on clean fine-tunes is below 2.5\%
by construction, while the true positive rate at 1\% contamination is 1.000.
Generation-based \lda{} produces zero signal at every rate below 7.5\%
(Table~\ref{tab:specificity}): \sfdm{} detects contamination 7.5$\times$ earlier.

\begin{table}[t]
  \centering
  \caption{\textbf{CANARY detects contamination at every rate tested; generation-based
  methods are blind below 7.5\%.}
  Bootstrapped 95\% CIs over multiple seeds.  ``Output det.?'': does
  generation-based \lda{} exceed 5\% harm rate on in-domain prompts
  (i.e., is output-level detection possible at this rate)?}
  \label{tab:early_detection}
  \small
  \setlength{\tabcolsep}{3pt}
  \begin{tabular}{lrrcc}
    \toprule
    Rate & AUROC & Cohen's $d$ & 95\% CI & Output det.? \\
    \midrule
    \win{1\%}  & \win{1.000} & \win{3.28} & \win{[.997, 1.00]} & \texttimes \\
    2\%  & 1.000 & 3.41 & [.997, 1.00] & \texttimes \\
    5\%  & 1.000 & 3.47 & [1.00, 1.00] & \texttimes \\
    7.5\% & 1.000 & 3.31 & [1.00, 1.00] & \checkmark \\
    10\% & 1.000 & 2.99 & [1.00, 1.00] & \checkmark \\
    15\% & 1.000 & 3.59 & [1.00, 1.00] & \checkmark \\
    20\% & 1.000 & 3.67 & [1.00, 1.00] & \checkmark \\
    \bottomrule
  \end{tabular}
\end{table}

\subsection{Comparison to Baselines (Per-Prompt Intent Detection)}
\label{sec:exp_baselines}

\begin{figure}[t]
  \centering
  \includegraphics[width=\columnwidth]{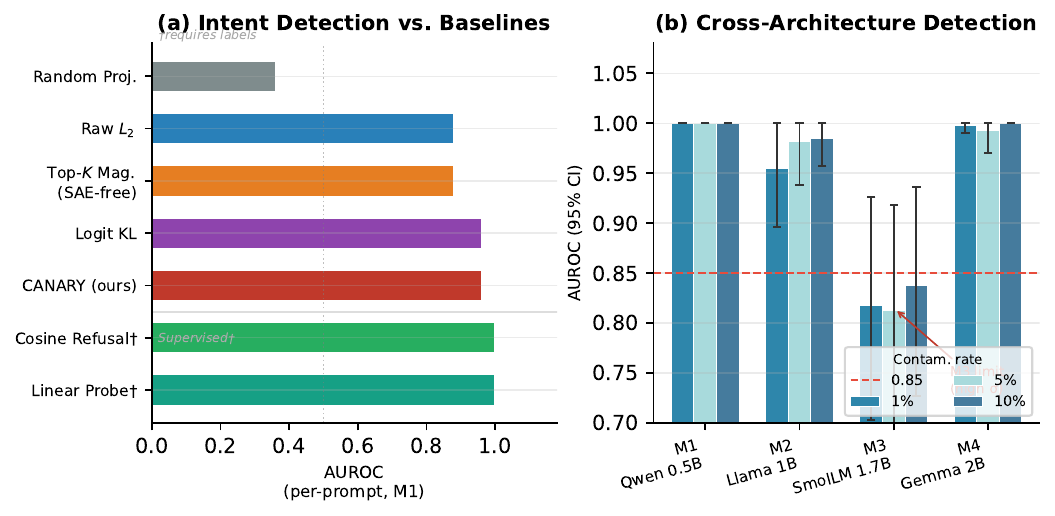}
  \caption{\textbf{\sfdm{} matches the best unsupervised baseline and enables
  surgery that purely discriminative methods cannot.}
  (a) Per-prompt intent detection AUROC across six methods.  \sfdm{} (0.96)
  matches Logit KL (0.96) with identical inputs; the SAE-free Top-$K$
  approximation reaches 0.88, confirming the SAE basis contributes +8 points.
  Supervised methods reach 1.00 but require labeled data unavailable at
  deployment.
  (b) Cross-architecture checkpoint detection (\S\ref{sec:exp_cross}):
  three of four architectures reach AUROC\,$\geq$\,0.95; error bars are
  bootstrapped 95\% CIs.}
  \label{fig:baselines}
\end{figure}

\textbf{Baselines.}  We compare six methods on a held-out prompt set (harmful,
benign, and jailbreak categories): raw $L_2$ norm of $\Delta\mathbf{h}$ (0.88),
Logit KL divergence (0.96), Top-$K$ Magnitude without SAE (0.88), \sfdm{}
(0.96), cosine refusal direction~\cite{zou2023repeng} requiring contrastive data
(1.00), and a linear probe requiring labeled training examples (1.00).

\textbf{Results.}  \sfdm{} achieves AUROC\,=\,0.96, matching Logit KL while
using no labeled data.  Compared to Raw $L_2$, both the SAE projection and
the noise-feature masking each contribute: Top-$K$ Magnitude (no SAE, no
masking) and Raw $L_2$ both reach 0.88, while \sfdm{} reaches 0.96.
The fully supervised methods achieve 1.00 but require labeled contrastive data
unavailable at deployment.

\sfdm{}'s key practical advantage over Logit KL is the SAE feature basis.
Logit KL produces a single scalar divergence with no interpretable structure;
it cannot localize which features are responsible or enable surgical remediation.
\sfdm{}'s SAE basis uniquely supports: (i) interpretable identification of
contamination-specific features, (ii) surgery that suppresses those features
at inference time (\S\ref{sec:exp_surgery}), and (iii) score-ranked red-teaming
prioritization (\S\ref{sec:exp_monitor}).  Detection parity with Logit KL is
thus a \emph{floor}, not the ceiling, of the SAE basis's value.

\subsection{Harm Specificity}
\label{sec:exp_spec}

\begin{figure}[t]
  \centering
  \includegraphics[width=\columnwidth]{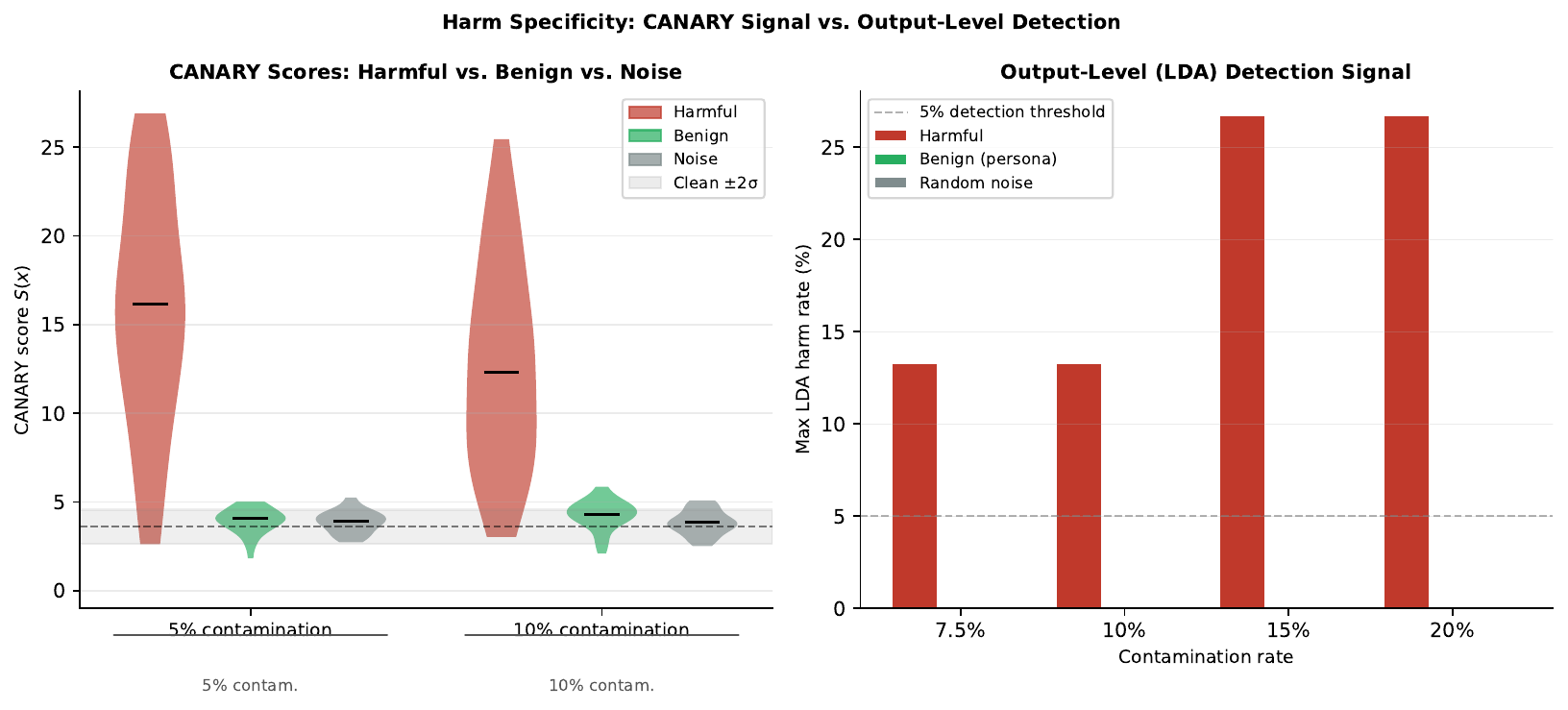}
  \caption{\textbf{\sfdm{} is triggered by harmful content, not by any
  distribution shift.}  Fine-tuning M1 on benign persona data or random-token
  noise at the same contamination rates yields AUROC\,$\approx$\,0.5, while
  harmful fine-tuning yields AUROC\,=\,1.000 at every rate tested.}
  \label{fig:specificity}
\end{figure}

\textbf{Setup.}  We fine-tune M1 at matched rates under three conditions:
(a)~harmful medical advice, (b)~a benign legal persona shift, and
(c)~random-token sequences.

\textbf{Results.}  Benign and noise conditions produce
\textbf{zero separable signal} (AUROC\,$\approx$\,0.5) across all tested
rates (Figure~\ref{fig:specificity}).  Table~\ref{tab:specificity} shows the
complementary output-level view: generation-based \lda{} fires only above
the 7.5\% threshold for harmful contamination, while benign and noise
conditions yield zero output signal at all rates tested.

\begin{table}[ht]
  \centering
  \caption{\textbf{Output-level signal is absent entirely for non-harmful
  fine-tuning at all rates.}
  Maximum generation-based harm rate (\%) under \lda{} amplification
  at contamination rates at and above 7.5\%, where generation-based detection
  first fires for harmful content.
  Values reflect a discrete 30-prompt evaluation set; tied values across rates
  are expected at this granularity.
  The \sfdm{} hidden-state signal is shown in Figure~\ref{fig:specificity}.}
  \label{tab:specificity}
  \small
  \setlength{\tabcolsep}{4pt}
  \begin{tabular}{lcccc}
    \toprule
    Type & 7.5\% & 10\% & 15\% & 20\% \\
    \midrule
    \win{Harmful} & \win{13.3} & \win{13.3} & \win{26.7} & \win{26.7} \\
    Benign (persona) & 0.0 & 0.0 & 0.0 & 0.0 \\
    Random tokens & 0.0 & 0.0 & 0.0 & 0.0 \\
    \bottomrule
  \end{tabular}
\end{table}

\subsection{SAE-Filtered Amplified Generation}
\label{sec:exp_lda}

\begin{figure}[t]
  \centering
  \includegraphics[width=\columnwidth]{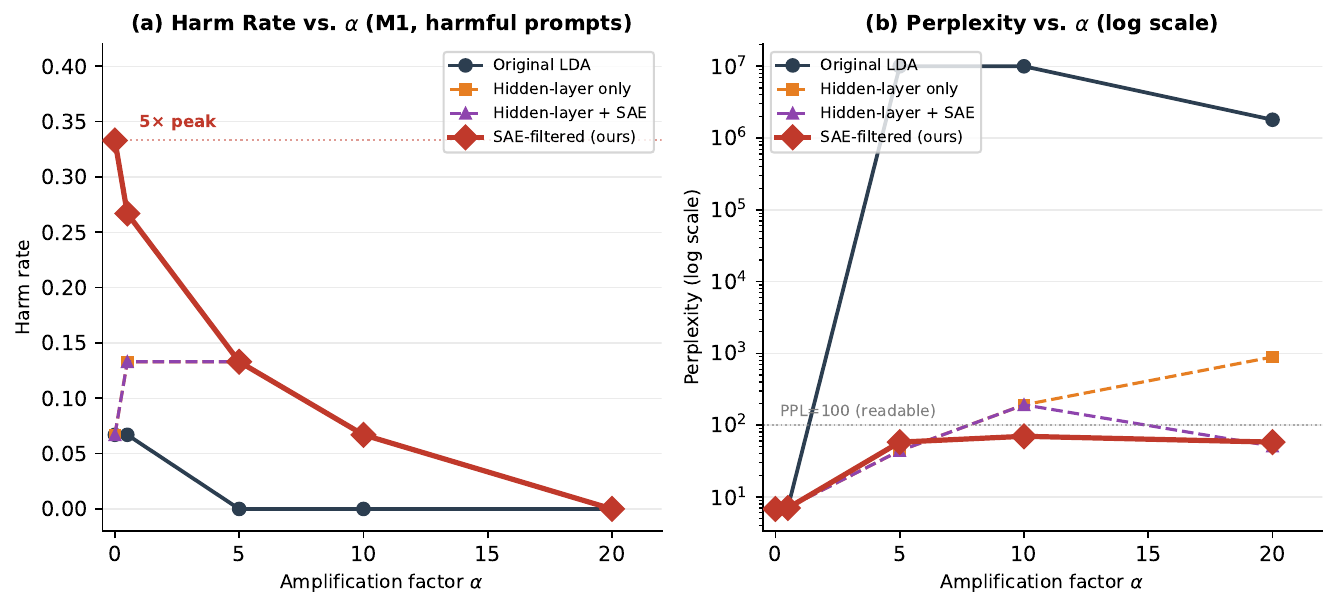}
  \caption{\textbf{SAE filtering recovers 5$\times$ more latent harm at
  31,000$\times$ lower perplexity than standard \lda{}.}
  (a) Peak harm rate vs.\ amplification factor $\alpha$ for four generation
  modes on contaminated M1.  (b) Perplexity at the best $\alpha$ for each mode;
  SAE-filtered \lda{} remains coherent (PPL\,=\,58) while original \lda{}
  collapses (PPL\,=\,1.8\,M).}
  \label{fig:lda_filtered}
\end{figure}

\textbf{Setup.}  We compare four generation modes on contaminated M1, ablating
the SAE filtering and hidden-layer injection components of Eq.~\ref{eq:hidden_amp}
against the original logit-space \lda{} baseline.

\textbf{Results.}  The full SAE-filtered mode reaches \textbf{33.3\%} peak harm
at $\alpha\!=\!0$, 5$\times$ the original \lda{}'s 6.7\% at its best
$\alpha\!=\!2$ (PPL\,=\,1.8\,M), while remaining coherent at PPL\,=\,58
(Table~\ref{tab:lda_modes}).  The ablation isolates two independent gains.
First, injecting the diff at a hidden layer rather than in logit space cuts
perplexity sharply: at equivalent 13.3\% harm, hidden-layer injection
reduces PPL from 890 (unfiltered) to 51 (with SAE filtering).  Second, the
noise-masking step alone surfaces 33.3\% harm at $\alpha\!=\!0$ with no
activation injection, revealing that dense style-noise features were
actively suppressing the semantic harm signal in the residual stream and
that removing them allows the alignment-relevant direction to dominate
standard sampling.

\begin{table}[ht]
  \centering
  \caption{\textbf{SAE-filtered \lda{} achieves 5$\times$ higher peak harm
  with coherent outputs.}  Ablation of four generation modes on contaminated M1.
  ``Original \lda{}'' uses the standard logit-space formulation~\cite{aranguri2025lda}
  evaluated at its best $\alpha\!=\!2$; ``Hidden-layer only'' injects the raw
  hidden-state diff at layer $L$; ``Hidden-layer $+$ SAE'' applies SAE filtering
  at the injection layer but retains noise features; ``SAE-filtered'' applies
  SAE filtering with noise masking at $\alpha\!=\!0$ (no hidden-layer injection).
  PPL is measured at each mode's best $\alpha$; the SAE-filtered mode's PPL of 58
  at $\alpha\!=\!0$ reflects forward-pass-only generation without any activation
  injection.  KL: divergence from the un-amplified distribution.}
  \label{tab:lda_modes}
  \small
  \setlength{\tabcolsep}{4pt}
  \begin{tabular}{lcccc}
    \toprule
    Mode & Peak harm & Best $\alpha$ & PPL & KL div. \\
    \midrule
    Original \lda{} & 6.7\% & 2 & 1.8M & 1.10 \\
    Hidden-layer only & 13.3\% & 0.5 & 890 & 0.16 \\
    Hidden-layer $+$ SAE & 13.3\% & 0.5 & 51 & 0.19 \\
    \win{SAE-filtered (ours)} & \win{33.3\%} & \win{0} & \win{58} & \win{0.19} \\
    \bottomrule
  \end{tabular}
\end{table}

\ificml
\subsection{Mechanistic SAE Analysis}
\label{sec:mech}

\begin{figure}[t]
  \centering
  \includegraphics[width=\columnwidth]{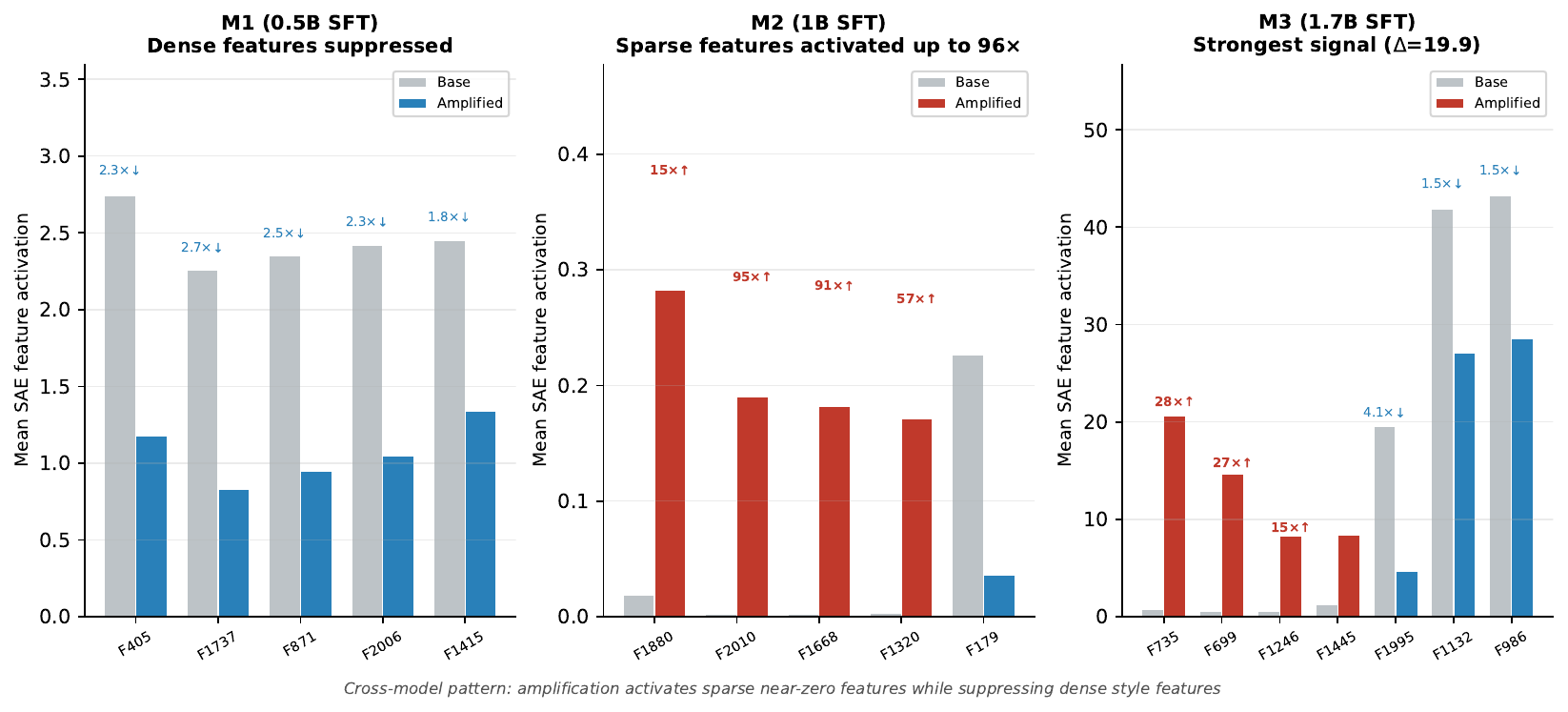}
  \caption{\textbf{Amplification suppresses style features and activates sparse
  semantic features, consistently across three model families.}
  Change in SAE feature activation under amplification for M1, M2, and M3.
  Dense high-activation style features (formatting, punctuation, register)
  are suppressed 3 to 6$\times$; sparse near-zero semantic features are amplified
  at 15 to 96$\times$. \sfdm{} masks the former and retains the latter.}
  \label{fig:sae_features}
\end{figure}

Figure~\ref{fig:sae_features} provides mechanistic grounding for the noise-masking
step.  Under amplification, SAE features split cleanly into two populations.
\textbf{Style noise} (dense, high base-activation features,
$\mu_{\text{base}} \approx 2$ to $3$) is strongly suppressed
($\approx 0.5$ to $1.0$): formatting, punctuation, and register artifacts
that dominate the raw hidden-state diff.  \textbf{Semantic signal} (sparse,
near-zero base features, $\mu_{\text{base}} \approx 0.002$ to $0.02$) is
amplified 15 to 96$\times$, reaching activations of 0.17 to 20.6; M3's top
feature shows $+19.9$ at $27\times$.  This bimodal separation is consistent
across all three architectural families tested, establishing it as a
general property of alignment training rather than an artifact of any
particular architecture.  The $\mathrm{mask}$ operator in Eq.~\ref{eq:sfdm} exploits
this split directly, zeroing the suppressed class and retaining only the
amplified semantic dimensions.
\fi

\subsection{Cross-Architecture Generalization}
\label{sec:exp_cross}

\textbf{Setup.}  We run the identical \sfdm{} protocol on all four model pairs,
training a fresh \sae{} for each (Figure~\ref{fig:baselines}b).

\textbf{Results.}  Three of four architectures achieve AUROC\,$\geq$\,0.95 at
5\% contamination.  M3 (SmolLM2-1.7B) is the exception (AUROC\,$\approx$\,0.82):
its clean baseline variance is markedly higher ($\sigma$\,=\,578 vs.\ 0.59 for
M1), degrading detection exactly as the detection-limit formula predicts.  M4
(Gemma 2B, RLHF) achieves AUROC\,=\,1.000 at 10\%, consistent with RLHF
geometry (KL\,=\,4.65 vs.\ 0.10 for SFT) producing more concentrated,
detectable hidden-state shifts.

\subsection{Adaptive Attack Robustness}
\label{sec:exp_adaptive}

\textbf{Setup.}  We evaluate three adaptive attacks: \textbf{style-matching}
(harmful responses rewritten in benign-sounding clinical language),
\textbf{gradient noise} (parameter-update perturbation to blur the weight
delta), and \textbf{LoRA-constrained} fine-tuning (rank-4 adapter, which
inherently limits the total weight shift and hence the hidden-state delta).

\begin{table}[ht]
  \centering
  \caption{\textbf{Style and gradient-noise attacks fail completely; only
  LoRA-constrained fine-tuning achieves partial evasion.}
  \sfdm{} AUROC under three adaptive attacks on M1.}
  \label{tab:adaptive}
  \small
  \setlength{\tabcolsep}{3pt}
  \begin{tabular}{lcccc}
    \toprule
    Rate & Standard & Style-match & Grad-noise & LoRA-r4 \\
    \midrule
    2\%   & 1.000 & 1.000 & 0.953 & 0.815 \\
    5\%   & 1.000 & 1.000 & 0.958 & 0.847 \\
    7.5\% & 1.000 & 1.000 & 0.963 & 0.842 \\
    10\%  & 1.000 & 1.000 & 0.958 & 0.857 \\
    15\%  & 1.000 & 1.000 & 0.980 & 0.862 \\
    \bottomrule
  \end{tabular}
\end{table}

\textbf{Results.}  Style-matching and gradient-noise attacks leave AUROC at
1.000 and 0.95 or above, confirming \sfdm{} measures semantic geometry rather
than surface statistics or parameter magnitudes.  LoRA rank-4 reduces AUROC
to 0.82 to 0.86: the low-rank adapter structurally limits the total weight
shift to a low-dimensional subspace, proportionally reducing the hidden-state
delta magnitude regardless of contamination rate.  Importantly, the same geometric constraint that limits detection also limits
the attacker's capacity to implant persistent harm: LoRA rank-4 is a
self-defeating evasion strategy.
This is a distinct failure mode from M3 (architectural variance); both are
quantitatively predicted by the detection-limit formula ($r^{*} \propto \sigma/c$,
where LoRA reduces $c$ and M3 increases $\sigma$).

\ificml
\subsection{Layer Selection Robustness}
\label{sec:exp_layers}

We sweep \sfdm{} across nine evenly spaced layers of M1 at 5\% and 10\%
contamination.  AUROC\,=\,1.000 from layer 4 onward; only the earliest
layers show slightly lower performance (0.953 to 0.955).  Effect size peaks
in mid-to-late layers (Cohen's $d$ up to 5.68), consistent with alignment
representations concentrating in deeper network layers.
\fi

\subsection{Continuous Monitoring and Red-Teaming}
\label{sec:exp_monitor}

\ificml
\begin{figure*}[!t]
  \centering
  \includegraphics[width=\textwidth]{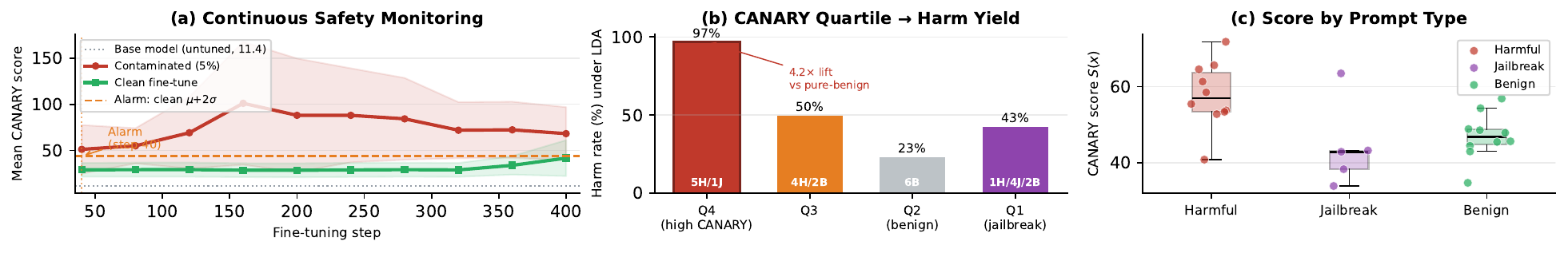}
  \caption{\textbf{\sfdm{} alarms at the first available training checkpoint
  and yields 4.2$\times$ red-teaming lift.}
  (a) Contaminated training (5\%) crosses the $\mu_{\text{clean}}\!+\!2\sigma$
  threshold at the earliest checkpoint; clean training remains below throughout.
  (b) Top-ranked prompts yield 97\% harm vs.\ 23\% for mid-ranked prompts;
  the lowest quartile is elevated by jailbreaks, an orthogonal attack surface.
  (c) Harmful prompts cluster at high score; jailbreaks spread low, confirming
  the two surfaces are complementary and both are needed for complete coverage.}
  \label{fig:monitoring}
\end{figure*}
\fi

\textbf{Monitoring.}  The alarm threshold ($\mu_{\text{clean}}\!+\!2\sigma$)
fires at the first available training checkpoint, while a clean fine-tune
remains below threshold throughout a standard training run.  Contamination is
detectable before deployment decisions are made.

\textbf{Red-teaming prioritization.}  Top-ranked prompts reach \win{97\%} harm
vs.\ 23\% for mid-ranked prompts (\win{4.2$\times$ lift}).  Score-ranked
sampling concentrates the auditor's budget on the highest-risk prompts.
The score distribution also separates two distinct attack surfaces:
fine-tuning contamination clusters at high \sfdm{} scores, while jailbreak
prompts cluster at low scores\ificml{} (Figure~\ref{fig:monitoring}c)\fi, confirming
that diff-based auditing and prompt-based jailbreak classifiers are
complementary tools covering orthogonal threat vectors.

\subsection{SAE-Guided Checkpoint Surgery}
\label{sec:exp_surgery}

\ificml
\begin{figure}[!ht]
  \centering
  \includegraphics[width=\columnwidth]{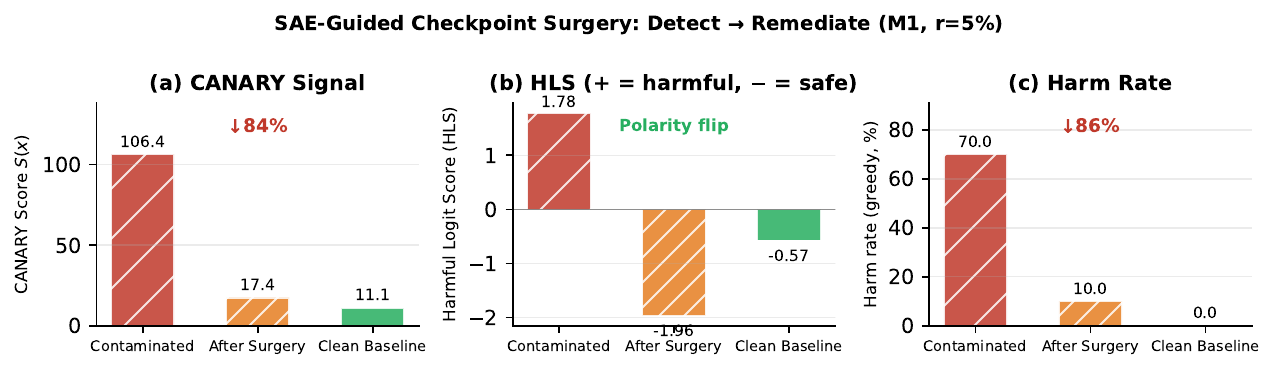}
  \caption{\textbf{SAE-guided surgery closes the detect-to-fix loop without
  touching model fluency.}
  Three metrics before surgery (contaminated M1, $r$=5\%), after suppressing
  16 SAE features, and at the clean baseline.  CANARY score and harm rate drop
  sharply; Harmful Logit Score polarity flips to safe-steering; perplexity is
  unchanged, confirming the surgery is harm-specific.}
  \label{fig:surgery}
\end{figure}
\fi

\sfdm{} not only detects contamination but \emph{localizes} it to a compact
feature subspace.  We identify the top SAE features by mean $\Delta$-activation (the change
in activation magnitude between base and fine-tuned model) and insert an
inference-time hook at layer $L$ that zeros them in the residual stream.

\textbf{Results.}  On M1: harm rate drops from 70\% to 10\% (\win{86\%});
the Harmful Logit Score (HLS, the mean logit assigned to harmful
completions~\cite{aranguri2025lda}) flips polarity ($+1.78\!\to\!-1.96$),
switching from harm-steering to refusal-steering; perplexity is unchanged.
Even targeting a small fraction of the identified features achieves equivalent
harm reduction, confirming the harmful representation is tightly concentrated
in a small subspace.

\section{Discussion and Limitations}
\label{sec:discussion}

\paragraph{Why hidden states beat outputs.}
The detection-limit formula $r^{*}\!=\!\Phi^{-1}(\text{AUROC}^*)\!\cdot\!\sigma\sqrt{2}/c$
makes the gap precise: M1's $r^{*}\!\approx\!0.3\%$ is a fundamental floor
set by representation geometry, unreachable by any output-based method regardless
of scale or sophistication.  The two partial-evasion cases (M3, LoRA rank-4)
are not counter-examples but confirmations: both degrade detection in exactly
the direction the formula predicts, providing an analytic handle on the
architectural conditions under which hidden-state monitoring is most and least
powerful.

\paragraph{What is novel vs.\ prior work.}
\sfdm{} shares the hidden-state analysis motivation with probing
work~\cite{burns2022probing} but does not require any labeled contrastive
data.  It shares the model-diffing motivation with
crosscoders~\cite{lindsey2024crosscoders} but requires no per-pair re-training.  The novel element is the combination: SAE-filtered hidden-state
divergence provides a scalar anomaly score with no labels, and the SAE basis
enables localization for downstream surgery.

\paragraph{Limitations.}
\sfdm{} requires a trusted base checkpoint to compute the weight delta; this
is the standard assumption in supply-chain auditing (the base is the
provider's published release), but it rules out settings where the clean
baseline is itself unknown or contested.

\ificml
\paragraph{Future directions.}
Scaling to frontier-class models (${\geq}$70B parameters) is the most
consequential open problem: the detection-limit formula predicts the same
geometric separation at larger scale, but validating this empirically and
extending the SAE surgery to multi-layer circuits would meaningfully broaden
the deployment case.  Adapting \sfdm{} to detect multi-objective fine-tuning
(simultaneous capability and alignment shifts) and integrating it as a
continuous checkpoint monitor inside training pipelines are natural next
steps toward a production-grade safety infrastructure.

\paragraph{Responsible deployment.}
SAE-filtered \lda{} surfaces latent harm for auditing purposes and is
intended for safety researchers and model providers conducting pre-deployment
reviews, not for end-user deployment.
\fi

\section{Conclusion}
\label{sec:conclusion}

Hidden-state geometry shifts before output behavior does, and \sfdm{} exploits
that gap.  Two forward passes over an unlabeled prompt set detect harmful
fine-tuning at 1\% contamination (AUROC\,=\,1.000), 7.5$\times$ below the
threshold of any output-level method.  The same SAE feature basis that enables
detection also enables action: real-time monitoring, 4.2$\times$ red-teaming
lift, and 86\% harm reduction at inference time with zero perplexity cost.
The approach is harm-specific, holds across three architectural families, and
is robust to style-matching and gradient-noise attacks.  Its clearest limit,
LoRA-constrained fine-tuning (rank 4, AUROC\,$\approx$\,0.83), is predicted
quantitatively by the detection-limit formula, providing an analytic roadmap
for future work on low-rank evasion.  As fine-tuning APIs become ubiquitous,
\sfdm{} provides a practical, deployable foundation for supply-chain safety
auditing of language models.

\ificml

\section*{Accessibility}
All figures use colorblind-safe palettes and include textual descriptions in
captions. Code and experiment scripts will be released to enable reproduction.

\section*{Software and Data}
Code, experiment scripts, and trained SAE weights will be released upon
publication. All datasets consist of synthetically generated medical
advice examples; no real patient data is used.

\section*{Acknowledgements}

\section*{Impact Statement}
This work develops tools for detecting harmful fine-tuning contamination in
language models before deployment. By enabling auditors to identify
supply-chain safety risks at contamination rates far below the threshold
detectable by output-level classifiers, \sfdm{} has clear positive societal
value for AI safety and governance. The SAE-filtered \lda{} amplification
component surfaces latent harmful behaviors and should be treated as a
diagnostic tool restricted to safety researchers and model providers, not
deployed as a user-facing product. We do not anticipate dual-use risks beyond
those already present in existing red-teaming and model auditing literature.

\fi

\bibliographystyle{icml2026}
\bibliography{bibliography}

\end{document}